\PassOptionsToPackage{export}{adjustbox}
\PassOptionsToPackage{hidelinks}{hyperref}
\PassOptionsToPackage{unicode}{hyperref}
\PassOptionsToPackage{naturalnames}{hyperref}

\documentclass[accepted]{uai2023} 

\usepackage[american]{babel}

\usepackage{natbib} 
\usepackage{mathtools} 
\usepackage{siunitx} 
\usepackage{booktabs} 
\usepackage{tikz} 

\usepackage{wrapfig}

\usepackage[ruled]{algorithm2e}

\usepackage{flafter}
\usepackage[utf8]{inputenc}
\usepackage{amsmath}
\usepackage{amssymb}
\usepackage{amsfonts}

\usepackage{graphicx}
\usepackage[colorinlistoftodos]{todonotes}

\usepackage{caption}
\usepackage{subcaption}
\definecolor{vibrant-blue}{RGB}{0,119,187}
\definecolor{vibrant-cyan}{RGB}{51,187,238}
\definecolor{vibrant-teal}{RGB}{0,153,136}
\definecolor{vibrant-orange}{RGB}{238,119,51}
\definecolor{vibrant-red}{RGB}{204,51,17}
\definecolor{vibrant-magenta}{RGB}{238,51,119}
\definecolor{vibrant-grey}{RGB}{187,187,187}

\definecolor{muted-indigo}{RGB}{51,34,136}
\definecolor{muted-cyan}{RGB}{136,204,238}
\definecolor{muted-teal}{RGB}{68,170,153}
\definecolor{muted-green}{RGB}{17,119,51}
\definecolor{muted-olive}{RGB}{153,153,51}
\definecolor{muted-sand}{RGB}{221,204,119}
\definecolor{muted-rose}{RGB}{221,102,119}
\definecolor{muted-wine}{RGB}{136,34,85}
\definecolor{muted-purple}{RGB}{170,68,153}

\usepackage[capitalize,nameinlink]{cleveref}
\crefalias{objective}{equation}
\crefname{objective}{Objective}{Objectives}
\Crefname{objective}{Objective}{Objectives}
\crefdefaultlabelformat{(#2\color{muted-indigo}\textbf{#1}#3)}

\DeclareCaptionFormat{captionsize}{\fontsize{8}{10}\selectfont#1#2#3}
\DeclareCaptionFormat{subcaptionsize}{\fontsize{7}{8}\selectfont#1#2#3}
\usepackage{glossaries-extra}
\setabbreviationstyle[long]{long}             
\setabbreviationstyle[long-noshort]{long-em}  
\newabbreviation[category=long-short]{nn}{NN}{neural network}
\newabbreviation[category=long-short]{mlp}{MLP}{multi-layer perceptron}
\newabbreviation[category=long-short,longplural={Gaussian processes}]{gp}{GP}{Gaussian process}
\newabbreviation[category=long-short]{smk}{SMK}{spectral mixture kernel}
\newabbreviation[category=long-short]{mk}{MK}{Mat\'{e}rn kernel}
\newabbreviation[category=long-short]{sgd}{SGD}{stochastic gradient descent}
\newabbreviation[category=long-short]{relu}{ReLU}{rectified linear unit}
\newabbreviation[category=long-short]{relu-nn}{rectifier NN}{rectifier neural network}
\newabbreviation[category=long-short]{gelu}{GELU}{Gaussian error linear unit}
\newabbreviation[category=long-short]{silu}{SiLU}{sigmoid-weighted linear unit}
\newabbreviation[category=long-short]{sin}{$\sin$}{sine}
\newabbreviation[category=long-short]{sin-nn}{sinusoidal NN}{sinusoidal neural network}
\newabbreviation[category=long-short]{tanh}{$\tanh$}{hyperbolic tangent}
\DeclareMathOperator{\erf}{erf}
\newabbreviation[category=long-short]{erf}{$\erf$}{Gauss error function}
\newabbreviation[category=long-short]{adam}{Adam}{adaptive momentum}
\newabbreviation[category=long-short]{ntk}{NTK}{neural tangent kernel}
\newabbreviation[category=long-short]{ard}{ARD}{automatic relevance determination}
\newabbreviation[category=long-short]{uci}{UCI}{UC Irvine Machine Learning Repository}
\newabbreviation[category=long-short]{mse}{MSE}{mean-squared error}
\newabbreviation[category=long-short]{ml}{ML}{marginal likelihood}
\newabbreviation[category=long-short]{mll}{MLL}{marginal log-likelihood}
\newabbreviation[category=long-short]{mcmc}{MCMC}{Markov chain Monte Carlo}

\glsdisablehyper

\newcommand{\X}{\mathbf{X}}
\newcommand{\y}{\mathbf{y}}

\newcommand{\modelfamily}{\mathfrak{F}}
\newcommand{\model}{\mathbf{g}}
\newcommand{\modelindex}{r}
\newcommand{\modelindexfinal}{R}
\newcommand{\datasetfamily}{\mathfrak{D}}
\newcommand{\dataset}{\mathcal{D}}
\newcommand{\datasetindex}{s}
\newcommand{\datasetindexfinal}{S}
\newcommand{\nbpairs}{T}

\newcommand{\eg}{\textit{e.g.,}~}
\newcommand{\ie}{\textit{i.e.,}~}

\newcommand{\parencite}{\citep}
\newcommand{\textcite}{\citet}

\title{Gaussian Process Surrogate Models for Neural Networks}

\author[1]{\href{mailto:<michaelyli@stanford.edu>?Subject=Your UAI 2023 paper}{Michael~Y.~Li}{}}
\author[2]{Erin~Grant}
\author[3]{Thomas~L.~Griffiths}
\affil[1]{%
    Department of Computer Science\\
    Stanford University\\
    Stanford, California, USA 
}
\affil[2]{%
    Gatsby Computational Neuroscience Unit\\
    University College London\\
    London, UK
}
\affil[3]{%
    Departments of Psychology and Computer Science\\
   Princeton, NJ, USA
  }
  
\begin{document}
\maketitle

\begin{abstract}
Not being able to understand and predict the behavior of deep learning systems makes it hard to decide what architecture and algorithm to use for a given problem.
In science and engineering, \emph{modeling} is a methodology used to understand complex systems whose internal processes are opaque.
Modeling replaces a complex system with a simpler, more interpretable surrogate.
Drawing inspiration from this, we construct a class of surrogate models for neural networks using Gaussian processes.
Rather than deriving kernels for infinite neural networks, we learn kernels empirically from the naturalistic behavior of finite neural networks.
We demonstrate our approach captures existing phenomena related to the spectral bias of neural networks, and then show that our surrogate models can be used to solve practical problems such as identifying which points most influence the behavior of specific neural networks and predicting which architectures and algorithms will generalize well for specific datasets.
\end{abstract}

\section{Introduction}
\label{sec:intro}

Deep learning systems are ubiquitous in machine learning but sometimes exhibit unpredictable and  undesirable behavior when deployed in real-world applications \citep{geirhos2020shortcut,d2020underspecification}.
This gap between idealized and real-world performance has driven calls for explainability, transparency, and interpretability of deep learning systems~\parencite{lipton2016mythos,doshi2017towards,samek2017explainable},
especially as these systems are more widely applied~\parencite{bommasani2021opportunities}.

Machine learning is not unique in seeking to understand a complex system whose inputs and outputs are observable but whose internal processes are opaque---this challenge occurs across the empirical sciences and engineering.
An explanatory tool that is foundational across these disciplines is that of \emph{modeling}, that is, representing a complex and opaque system with a simpler one that is more amenable to interpretation.\footnote{
  Though some architectural components of a deep learning system are commonly referred to as a \emph{model}---as in ``neural network model''---we use \emph{modeling} to refer to the methodology of idealizing a complex system as a simpler one.}
Modeling makes precise assumptions about how a system may operate while abstracting away details that are irrelevant for a given level of understanding or a given downstream use case.
These properties are valuable for a framework for understanding deep learning as they are in other scientific and engineering disciplines.

As the popularity of deep learning has grown, a number of proposals have been made for modeling these systems.
Numerous mathematical models of deep learning have been developed~\parencite{PDLT-2022},
and some surprising phenomena,
such as adversarial examples~\parencite{szegedy2014intriguing},
have been captured with a mathematical analysis~\parencite{ilyas2019adversarial}.
However, existing mathematical models 
are unable to capture the properties of machine learning systems as applied in practice~\parencite{nakkiran2021towards}.
Beyond mathematical models,
specific models have aimed to explain the predictions of machine learning systems on a per-example basis~\parencite{Ribeiro2016WhySI,koh2017understanding,zhou2022exsum,ilyas2022datamodels},
but these approaches are, by construction, only partial explanations of the end-to-end system.

In this paper, we pursue an alternative modeling approach that draws on two domains for inspiration.
In engineering design,
\emph{surrogate models}~\parencite{wang2006review} emulate the input-output behavior of a complex physical system, allowing practitioners to simulate effects that are consequential for design or analysis without relying on costly or otherwise prohibitive queries from the system itself. In cognitive science,
\emph{cognitive models}~\parencite{sun2008cambridge,mcclelland2009place} describe how unobservable mental processes such as memory or attention produce the range of people's observed behaviors.
Both domains abstract away internal details,
such as real-world constraints on a physical system or neural circuitry in the brain, instead treating the target process or system as a \emph{black box}.
At the same time, both surrogate and cognitive models are constructed to replicate the end-to-end behavior of the target system and thus are complete where localized explanations are not.

We explore an analogous approach to investigate deep learning systems by constructing \emph{Gaussian process surrogate models for neural networks}.
\Glspl{gp} are a natural choice, with appealing theoretical properties specific to the study of \glspl{nn};
namely, certain limiting cases of \gls{nn} architectures are realizable as \glspl{gp}~\parencite{neal1996priors,li2018learning,jacot2018neural,allen2019convergence,du2019gradient}.
However, in contrast to these theoretical approaches, we explore the scientific and practical utility of idealizing \glspl{nn} with \glspl{gp}
using a \emph{data-driven} approach to estimating the kernel functions of finite \glspl{nn}.
With this approach,
we capture a number of known phenomena, including a bias towards low frequencies and pathological behavior at initialization, in a cohesive framework.
Finally, we demonstrate the practical benefits of this framework by identifying points that strongly influence NN predictions and predicting
the generalization behavior of models in a \gls{nn} family.

\section{Background}
\label{sec:background}

In \textbf{surrogate modeling}, we approximate a complex function with a simpler surrogate model that emulates its input-output behavior.
Surrogate models have many applications:
In optimization, they are often used to approximate queries from expensive-to-evaluate functions~\parencite{snoekpracbayes,shahriari2016};
in other applications, surrogate models have been used to gain insight into large physical systems~\parencite{campsvallsrankingdrivers}.

\textbf{Cognitive models} have been used  by cognitive scientists since the 1950s to gain insight into the human mind~\parencite{newell1958elements}.
Bayesian models of cognition offer a way to describe the inductive biases of learning systems as a prior distribution~\parencite{griffiths2010probabilistic}.
As deep \glspl{nn} have become more prevalent, researchers have used methodologies from cognitive science to interrogate opaque models~\parencite{ritter2017cognitive,geirhos2018imagenet,hawkins2020investigating}.
The success of these efforts suggests that other methods from cognitive science---namely, cognitive modeling---may be  applicable to machine learning systems.

\textbf{Gaussian processes}~\parencite[GPs;][]{rasmussenbook} are probabilistic models that specify a distribution over functions.
A \gls{gp} models any \emph{finite} set of $N$ observations as a multivariate Gaussian distribution on~$\mathbb{R}^N$,
where the $n$th point is interpreted as the function value, $f(\mathbf{x}_n)$, at the input point $\mathbf{x}_n$.
\Gls{gp}s are fully characterized by a mean function $m(\mathbf{x})$, usually taken to be degenerate as $m(\mathbf{x}) = \mathbf{0}, \forall \mathbf{x}$, and a positive-definite kernel function $k(\mathbf{x},\ \mathbf{x'})$ that gives the covariance between~$f(\mathbf{x})$ and~$f(\mathbf{x'})$ as a function of~$\mathbf{x}$ and~$\mathbf{x'}$.

Formally,
let $\mathbf{X}$ be a matrix of inputs and $\mathbf{y}$ be a vector of output responses.
Due to the marginalization properties of the Gaussian distribution, the posterior predictive distribution of a \gls{gp} for a new input $\mathbf{x_{*}}$, conditioned on dataset~${\dataset = \{ \mathbf{X}, \mathbf{y}\}}$ and assuming centered Gaussian observation noise with variance $\sigma^2$, is Gaussian with closed-form expressions for the mean and variance:
\begin{align}
  \mathbb{E}[f(\mathbf{x_{*}}) \mid \dataset] & = m(\mathbf{x_{*}}) + \mathbf{k_{*}}^{T}(\mathbf{K} + \sigma^2\mathbf{I})^{-1}(\mathbf{y}-m(\mathbf{x}))
  \label{eq:pp-mean}
  \\
  \mathbb{V}[f(\mathbf{x_{*}}) \mid \dataset] & =
  k(\mathbf{x_{*}}, \mathbf{x_{*}})- \mathbf{k_{*}}^{T}(\mathbf{K} + \sigma^2\mathbf{I})^{-1}\mathbf{k_{*}}
  \label{eq:pp-var}
\end{align}
where $\mathbf{K}$ is the ${N\times N}$ Gram matrix of pairwise covariances, 
$k(\mathbf{x}_i, \mathbf{x}_j)$,
and~${\mathbf{k_{*}} = [k(\mathbf{x}_1, \mathbf{x_{*}}), \ldots, k(\mathbf{x}_N, \mathbf{x_{*}})]^{T}}$.

The kernel function $k$ specifies the prior on what kind of functions might be represented in observed data, for example, expressing expected smoothness or periodicity.
Parametric kernels have hyperparameters $\theta$ that affect this prior and thus the posterior predictive.
These kernel hyperparameters can be adapted to the properties of a dataset, thus defining a prior over functions that is appropriate for that context.
\Gls{gp} kernel hyperparameters are typically learned via gradient-based optimization to maximize the \gls{gp} marginal likelihood, $p(\mathbf{y} \mid\mathbf{X})$.
Again due to properties of the \gls{gp}, this marginal likelihood has the closed-form expression:
\begin{align}
  \log p(\mathbf{y} \mid\mathbf{X}) =& -\frac{1}{2}\mathbf{y}^T\left(\mathbf{K}_{\theta}+\sigma^2_n I\right)^{-1}\mathbf{y} \nonumber \\ &- \frac{1}{2}\log|\mathbf{K}_{\theta} +\sigma^2_n I|-\frac{n}{2}\log2\pi~.
  \label{eq:marginal}
\end{align}
We write the Gram matrix as $\mathbf{K}_{\theta}$ to indicate that it depends on kernel hyperparameters via a particular parameterization.
In this work, we make use of two kernel parameterizations:
the \textbf{\acrlong{mk}}~\parencite[\acrshort{mk};][]{matern1960spatial} and the \textbf{\acrlong{smk}}~\parencite[\acrshort{smk};][]{wilson2013kernels}.
\glsunset{smk}\glsunset{mk}
Specifically,
following \textcite{snoekpracbayes}, we use the \gls{ard} $5/2$ \gls{mk}, given by:
\begin{align}
  k(\mathbf{x}, \mathbf{x'})
  = & \ \theta_0 \left(1 + \sqrt{5 {r^2(\mathbf{x}, \mathbf{x'})}} + \tfrac{5}{3} r^2(\mathbf{x}, \mathbf{x'})\right) \nonumber \\ 
  & \qquad \qquad \ \ \  \exp\left\{ -5\sqrt{5 r^2(\mathbf{x}, \mathbf{x'})} \right\}
  \label{eq:matern_kernel}
\end{align}
where $r^2(\mathbf{x}, \mathbf{x'}) = \sum_{d = 1}^{D} (x_d - x'_d)^2/\theta_d^2$ and
each $\theta_d$ is a lengthscale parameter capturing smoothness along dimension $d$.
The \gls{smk} is derived by modeling the spectral density associated with a kernel as a scale-location mixture of Gaussians~\parencite{wilson2013kernels}, giving:
\begin{align}
  k(\tau) = \sum_{q=1}^{Q} w_q \cos\left(2\pi^{2}\tau^{T}\boldsymbol{\mu_q}\right)\prod_{p=1}^{P} \exp\left\{-2\pi^2\tau_p^2v_q^{(p)}\right\}.
  \label{eq:smk_kernel}
\end{align}
Here, $k(\tau)$ gives the covariance between function values $f(\mathbf{x})$ and $f(\mathbf{x'})$ whose corresponding $P$- dimensional input values $\mathbf{x}$ and $\mathbf{x'}$ are a distance $\tau$ apart.
For a $Q$-component spectral mixture,
$w=\lbrace w_{i} \rbrace_{i=1}^Q$ are scalar mixture weights, 
the $q$-th component has mean vector $\boldsymbol{\mu_{q}} = (\mu^{(1)}_q, \ldots, \mu^{(P)}_q)$
and covariance matrix $\text{diag}(v_{q}^{(1)}, \ldots, v_{q}^{(P)})$.
See Section A.3 of the supplement for details on how the hyperparameters of the \gls{mk} and the \gls{smk} control the priors on functions.

\begin{figure}[!t]
  \renewcommand{\arraystretch}{2}
  \centering
  \captionsetup[subfigure]{labelformat=empty}
  \begin{subfigure}[b]{0.22\textwidth}
    \centering
    \scalebox{0.7}{
    \begin{tabular}{ccccc}
      $\mathbf{X}^1$
       & $\rightarrow$
       & \includegraphics[valign=m,width=30pt]{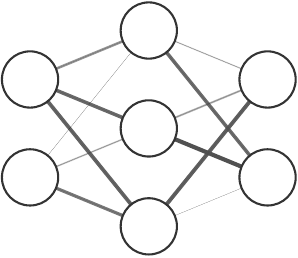}
       & $\rightarrow$
       & $\model_1(\mathbf{X}^1)$
      \\
       &                                                                 & \vdots &  & \\
      $\mathbf{X}^\datasetindexfinal$
       & $\rightarrow$
       & \includegraphics[valign=m,width=30pt]{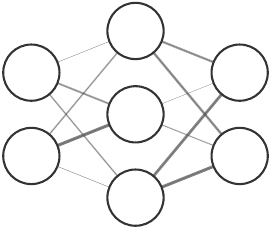}
       & $\rightarrow$
       & $\model_\modelindexfinal(\mathbf{X}^\datasetindexfinal)$                      \\
    \end{tabular}
    }

    \caption{\textbf{Step 1}:
      Collect predictions across models $\model_\modelindex$
      and across target functions (datasets) $\mathbf{X}_\datasetindex$.
    }
  \end{subfigure}
  \hfill
  \begin{subfigure}[b]{0.2\textwidth}
    \centering

    \includegraphics[width=80pt]{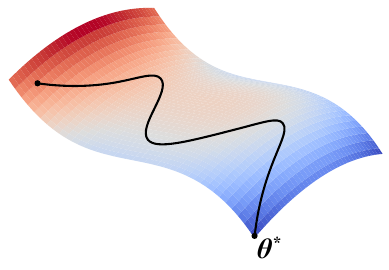}
    \caption{\textbf{Step 2}: Fit \acrlong{gp} hyperparameters $\theta$ to
      the aggregate predictions
      via \text{\cref{eq:objective}}.
    }
  \end{subfigure}
  \vspace{10pt}

  \begin{subfigure}[b]{0.48\textwidth}
    \centering
    \includegraphics[width=0.7\textwidth]{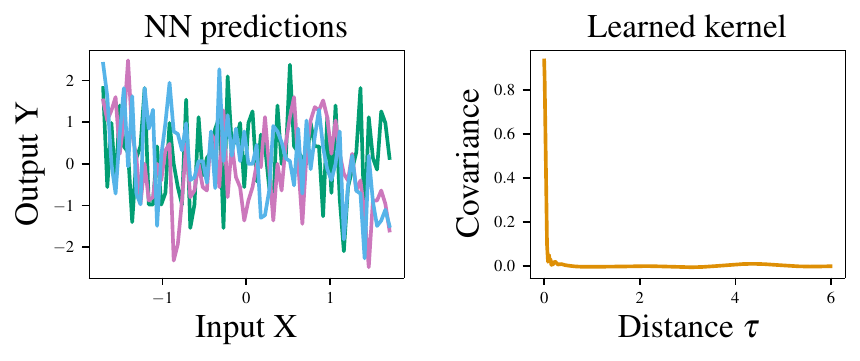}
    \caption{\textbf{Step 3}:
      Analyze the kernel learned from the aggregate predictions.
      Here, the learned surrogate kernel reveals the quickly varying behavior of particular \acrlong{nn}s.
    }
  \end{subfigure}
  \caption{
    \textbf{Outline of the surrogate modeling approach.}
    We learn a \acrlong{gp} surrogate model for a neural network family applied to a task family
    by learning kernel hyperparameters from neural network predictions across datasets.
    The learned kernel provides insight into the properties of the neural network family; e.g., biases towards particular frequencies~(\cref{sec:spectral-bias}),
    or expected generalization behavior~(\cref{sec:gen-gap}).
  }
  \label{fig:approach}
\end{figure}

\section{Learning a GP surrogate model from NN predictions}
\label{sec:method}
\begin{algorithm}[t]

  \SetKwInOut{HyperParameters}{hyperparameters}

  \HyperParameters{
    model family $\modelfamily$, \\
    \ dataset family $\datasetfamily$, \\
    \ model-dataset count $N$, \\
    \ GP parameterization $\theta$ \\
  }
  \tcp{Step 1 in \cref{fig:approach}}
  \For{$n \in 1 \hdots N$}{
    Sample a model, $\model_{\modelindex_n} \sim \text{Unif}(\modelfamily)$\\
    Sample a dataset, $\dataset_{\datasetindex_n} \sim \text{Unif}(\datasetfamily)$\\
    Train the model, $\model^\text{fit}_{\modelindex_n} \gets $ train$(\model_{\modelindex_n}, \dataset^\text{train}_{\datasetindex_n})$\\
    Evaluate $\model^\text{fit}_{\modelindex_n}(\dataset^\text{eval}_{\datasetindex_n})$\\
  }
  \tcp{Step 2 in \cref{fig:approach}}
  Optimize \cref{eq:objective} for $\theta^*$ \\
  \tcp{Step 3 in \cref{fig:approach}}
  Analyze $\theta^*$ via $P_{\theta^*}$\\
  \caption{Training and evaluation of a GP surrogate.}
  \label{alg:algorithm}
\end{algorithm}

In this section, we detail the goals and approach of the surrogate modeling framework. 
In brief, our approach involves collecting neural network predictions across a set of initializations and datasets and estimating \gls{gp} kernel hyperparameters from these predictions by maximizing the marginal likelihood across network-dataset pairs; see \cref{fig:approach}.

\subsection{Formal framework}
\label{sec:method-formal}
Our goal is to capture shared properties among a family of neural network models $\modelfamily$ as applied to a family of datasets $\datasetfamily$; untrained neural networks are the special case where $\datasetfamily$ is empty.
Here, a model family $\modelfamily$ is a set of neural networks $\{\model_0, \hdots, \model_\modelindexfinal\}$
that share in design choices (\eg architecture, training procedure, random initialization scheme) but differ in quantities that are randomized prior to or during training (\eg parameter initializations).
Similarly, a dataset family $\datasetfamily$ is a set of datasets
$\{\dataset_0, \hdots, \dataset_\datasetindexfinal\}$
that share some underlying structure
as in multi-task and meta-learning settings~\parencite{caruana1997multitask,hospedales2020meta}.
We consider supervised learning, in which each dataset consists of inputs and targets, $\dataset = (\X, \y)$.
Importantly, we fit surrogate model parameters $\theta$ to a \emph{behavioral dataset} of the model family evaluated on the dataset family, and not the ground truth datasets themselves.
By behavioral dataset, we mean a dataset of neural network predictions at some test inputs.
\paragraph{Data.}
We construct a component of the surrogate model training dataset as follows:
We sample a model index $\modelindex$ and a dataset index $\datasetindex$.
The corresponding dataset is split into a training set and an evaluation set,
$\dataset_\datasetindex =
  \dataset_\datasetindex^\text{train} \cup
  \dataset_\datasetindex^\text{eval}$.
The corresponding model $\model_\modelindex$ is fit the
training set
$\dataset_\datasetindex^\text{train} =
  (\X_\datasetindex^\text{train}, \y_\datasetindex^\text{train})$
according to the training procedure specified by the choice of model family $\modelfamily$, producing $\model_\modelindex^{\text{fit}}$.
We then collect the predictions of the trained model on the evaluation set,
$\model_\modelindex^\text{fit}(\X_\datasetindex^\text{eval})$,
to produce the component
$(\X_\datasetindex^\text{eval},
  \model_\modelindex^\text{fit}(\X_\datasetindex^\text{eval}))$
consisting of the \emph{ground truth inputs}
paired with the \emph{neural network behavioral targets} (\ie neural network predictions on ground truth inputs) from the evaluation set.
We aggregate the ground truth inputs and the neural network behavioral targets across pairs to produce the
\emph{surrogate model training dataset},
$\left(
  (\mathbf{X}^{\text{eval}}_{\datasetindex_1},\model_{\modelindex_1}^\text{fit}(\X^{\text{eval}}_{\datasetindex_1})),
  \hdots,
  (\mathbf{\X}^{\text{eval}}_{\datasetindex_\nbpairs},\model_{\modelindex_\nbpairs}^\text{fit}(\X^{\text{eval}}_{\datasetindex_\nbpairs}))
  \right)$.

\paragraph{Surrogate model.}
We fit the \gls{gp} using type-II maximum likelihood.
Let $P_{\theta}(\model^\text{fit}(\X^\text{eval}) \mid \X^\text{eval})$
be the \gls{gp} marginal likelihood of the dataset component
$\left(\mathbf{\X^{\text{eval}}},\model^\text{fit}(\X^{\text{eval}})\right)$
under a \gls{gp} with kernel hyperparameters $\theta$,
as given in~\cref{eq:marginal}.
We fit the surrogate model
jointly across model-and-task pairs in the surrogate model training dataset
by maximizing the marginal likelihood with respect to $\theta$:
\begin{flalign}
  \max_\theta \prod_{(\modelindex,\datasetindex)}
  P_{\theta}(\model^\text{fit}_\modelindex(\X^{\text{eval}}_{\datasetindex}) |
  \mathbf{X}^{\text{eval}}_\datasetindex)~.
  \label[objective]{eq:objective}
\end{flalign}
By optimizing~\cref{eq:objective},
we encourage the kernel hyperparameters $\theta$ to capture the implicit posterior distribution over functions
induced by the models in the family $\modelfamily$ as applied to the datasets in the family $\datasetfamily$. \Cref{alg:algorithm} gives the complete surrogate model training and evaluation process.

\paragraph{Motivating the framework.}
By estimating a posterior over functions for a neural network family, we aim to capture shared properties that determine the model family's behavior on data, \ie \textbf{the model family's \emph{inductive biases}}.
The inductive biases of neural networks (\eg invariances and equivariances, Markovian assumptions, compositionality) play an important role in their performance
by determining their extrapolation behavior~\parencite{mitchell1980need}. 

\Glspl{gp}, in particular, offer several advantages as surrogate models of \glspl{nn}. Firstly,
\glspl{gp}
\textbf{ are flexible models that are also often interpretable} in the sense that the learned hyperparameters can provide insights into properties of the
datasets on which they are trained~\parencite{wilson2013kernels}.
Secondly,
the use of \gls{gp} surrogate models is also motivated by
the \textbf{theoretical connections between \glspl{gp} and \glspl{nn}.}
\textcite{neal1996priors} showed that a prior over functions, implied by a prior over the weights of certain single-layer \glspl{mlp}, converges to a \gls{gp} as the \gls{mlp}'s width approaches infinity,
and recent works~\parencite{leedeep,matthews2018gaussian,novak2019bayesian,garriga-alonso2018deep,yang2019} have extended this correspondence to deep \glspl{mlp} and more modern \gls{nn} architectures.
The connections between \glspl{gp} and \glspl{nn} can provide insight because they transform the priors implicit in \glspl{nn} designs into explicit priors expressed through a \gls{gp}.
However, our strategy differs from this prior theoretical work that derives analytic kernels for limiting cases of NNs---we instead take an empirical approach by learning \gls{gp} kernels from the predictions of arbitrary classes of finite \glspl{nn}. Lastly, \glspl{gp} have \textbf{analytic} marginal likelihoods and leave-one-out predictive distributions, 
a property that we will exploit in \cref{sec:influence-functions-label,sec:gen-gap} to reduce costly computations.

\subsection{Demonstration: Comparing learned GP priors with NN priors}
\label{sec:learned-priors}
\begin{figure}[t]
  \centering
  \begin{subfigure}{0.95\textwidth}
    {\includegraphics[scale=0.28]{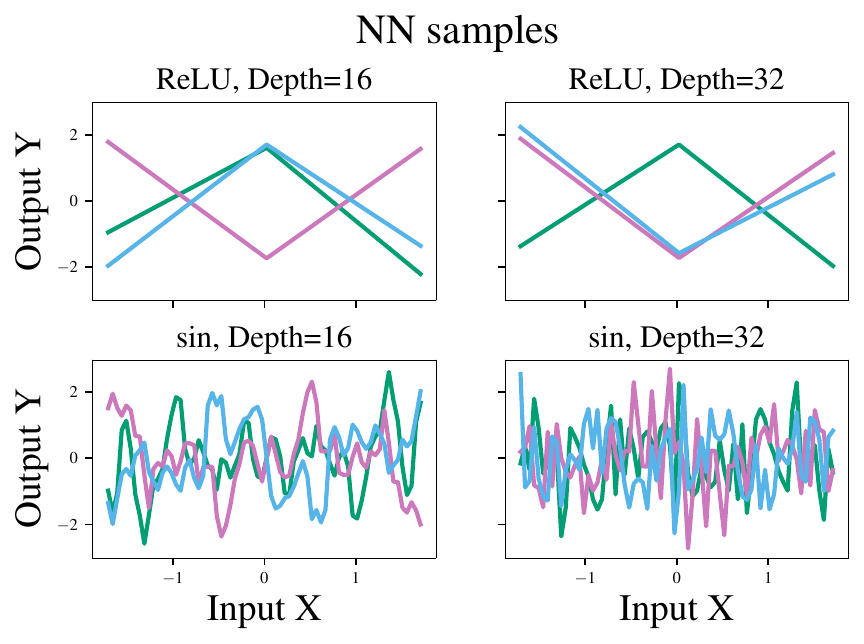}}
      {\includegraphics[scale=0.28]{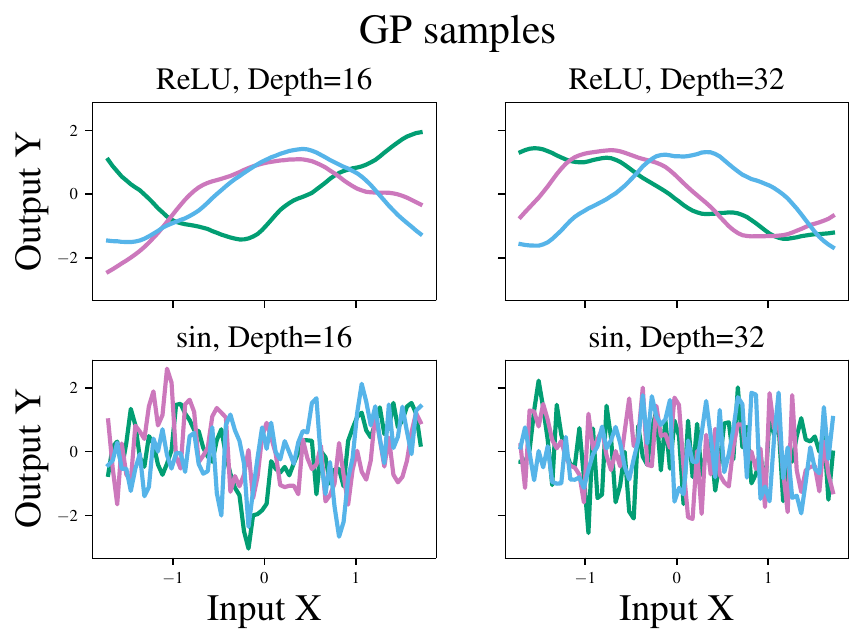}}
  \end{subfigure}
  \caption{
    \textbf{Demonstration: Comparing learned \gls{gp} priors with \gls{nn} priors.}
    Samples from \gls{gp} prior (\textbf{right}) with kernel hyperparameters inferred from the predictions of \gls{nn} families (\textbf{left}). 
    \gls{gp}s are flexible enough to capture properties of each \gls{nn} family; 
    for example, the samples from the learned \gls{gp} prior reflect the quickly varying behavior of the 32-layer sinusoidal NNs and the increasing-decreasing behavior of rectifier NNs.
  }
 \label{fig:nnvsgppriors}
\end{figure}

We briefly demonstrate the framework of \cref{sec:method-formal}
by verifying that \gls{gp} surrogates
learned from different \gls{nn} families exhibit meaningful variation in behavior.
\paragraph{\Gls{nn} hyperparameters.}
We consider ensembles of 50 randomly initialized (about zero with weight variance $\sigma_w^2=1.5$ and bias variance $\sigma_b^2=0.05$)
fully connected \glspl{nn} with \gls{relu} or \gls{sin} activations and 16 or 32 hidden layers of 128 hidden units each.
\paragraph{\Gls{gp} surrogate.}
For each ensemble, we learn the hyperparameters of a randomly initialized \gls{smk} with $Q=10$ mixture components by optimizing \cref{eq:objective} for 350 iterations
with batch gradient descent and the \gls{adam} optimizer \parencite{kingma:adam}  with learning rate $\eta = 0.1$.
We choose the kernel hyperparameters with highest objective value across three random initializations.

\paragraph{Results.}
We plot \gls{nn} predictions and samples from the learned \gls{gp} priors in \cref{fig:nnvsgppriors}.
The learned \gls{gp} captures the periodicity of the \glspl{sin-nn},
and partially captures the increasing-decreasing behavior of \glspl{relu-nn} about a cusp;
though, due to the \gls{smk} parameterization, it cannot capture the cusp.
The \gls{gp} also captures differences in depths for the \glspl{sin-nn}: The \gls{gp} prior samples for the 32-layer networks are quickly varying, indicating shorter learned lengthscales.
Taken together, the results of this demonstration show that \gls{gp} surrogates can capture
certain \gls{nn} behavior.
\begin{figure}[t]
  {\includegraphics[scale=0.64]{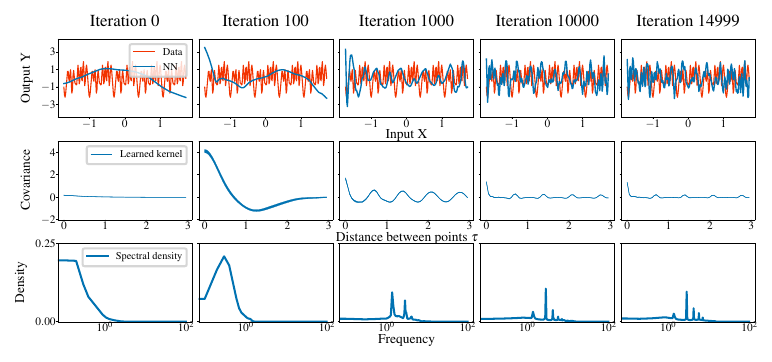}}  
  \caption{\textbf{Capturing spectral bias in neural networks.}
    (\textbf{Top}) Neural network predictions as training progresses on the sum-of-sines target function described in \cref{sec:spectral-bias}.
    (\textbf{Middle}) Spectral mixture kernel fit to neural network predictions as training progresses and
    (\textbf{Bottom})
    corresponding spectral density.
    The kernel reveals a spectral bias for this neural network family: the range of spectral frequencies expressed in the kernel increases with the number of iterations of training.
  }
  \label{fig:varysinex}
\end{figure}

\section{Experiments}
\label{sec:experiments}
We provide a series of demonstrations of the value of the approach of \cref{sec:method}.
Each experiment aims to investigate the properties of one or more neural network families, specified by \textbf{\glsfirst{nn} hyperparameters},
as evaluated on one or more dataset families, parameterized as \textbf{target functions},
by analyzing the corresponding \textbf{\glsfirst{gp} surrogate model}.
In \cref{sec:spectral-bias} and \cref{sec:random-init-prior}, we capture previously established \gls{nn} phenomena, 
in \cref{sec:influence-functions-label}, we identify influential points for \glspl{nn},
and 
in \cref{sec:gen-gap}, we predict \gls{nn} generalization behavior.

\subsection{Reproduction: Spectral bias}
\label{sec:spectral-bias}
\textcite{rahaman19a} demonstrated that deep \glspl{relu-nn} exhibit \emph{spectral bias}, the preference to learn lower frequencies in the target function before higher frequencies.
To demonstrate this, the authors studied the Fourier spectrum of \glspl{relu-nn} fit to a sum of sinusoidal functions of varying frequencies.
We take an alternative approach, learning kernels from \gls{nn} predictions at various stages of training. We show that these learned kernels capture the spectral bias.

\paragraph{\Gls{nn} hyperparameters.}
As in \textcite{rahaman19a},
we train an ensemble of 20 NNs with 6 hidden layers of 256 units and \gls{relu} activations using full-batch gradient descent with \gls{adam} and a learning rate of $\eta = 3 \times 10^{-4}$.

\paragraph{Target function.}
We consider 20 target functions which are sums of sine functions with frequencies in $(5, 10, \ldots, 45, 50)$ and phases drawn from $U(0, 2\pi)$, evaluated at $200$ points evenly spaced between $[0, 1]$, as in \textcite{rahaman19a}.

\paragraph{\Gls{gp} surrogate.}
We learn the parameters of a \glsfirst{smk} with $Q=10$ mixture components
by optimizing \cref{eq:objective} with \gls{adam} for $350$ iterations with a learning rate of $\eta = 0.1$ \parencite{kingma:adam}.
Since the marginal likelihood of the \gls{smk} is multi-modal in its frequency parameters, we repeat this optimization for three different random initializations of the kernel parameters and choose the hyperparameters with the largest marginal likelihood value (the value of \cref{eq:objective}).
Following GPyTorch’s initialization strategy, we draw the inverse lengthscale $v_i$ from a truncated Gaussian distribution with variance set to the maximum distance between two points in the dataset and truncation value set to the maximum distance between points \parencite{NEURIPS2018_gpytorch}. 
We set the signal variances $w$ to the variance of the target function values divided by the number of mixture components.
The frequency hyperparameters of the \gls{smk} are sometimes initialized by sampling from a uniform distribution whose upper limit is the Nyquist frequency \parencite{wilson2013kernels};
since this target function's largest frequency is smaller than the Nyquist frequency, we instead set a smaller frequency.

\paragraph{Results.}
\cref{fig:varysinex} displays the average \gls{nn} prediction across the ensemble, the average learned kernel and standard error at different iterations of \gls{nn} training computed across the ensemble and ten random initializations of the \gls{smk}, and the corresponding average spectral density.

The kernel function, which is given in \cref{eq:smk_kernel}, reflects how the similarity between function values varies with the distance between their input points.\footnote{Since the \gls{smk} is a stationary covariance function, we graph against the distance between input points}
The structure of the learned kernel reflects the properties of the \gls{nn} family:
Initially, the learned kernel only captures low frequencies in the \gls{nn}'s predictions---seen in the long period of the kernel---consistent with the spectral bias of \textcite{rahaman19a}.
However, with more training, the periodicity of the surrogate kernel reflects both low and high frequencies. 
Consistent with this, the spectral density exhibits a wider range of frequencies. 

\begin{figure}[t]
  \begin{subfigure}[t]{0.48\textwidth}
\includegraphics[width=0.90\linewidth, center]{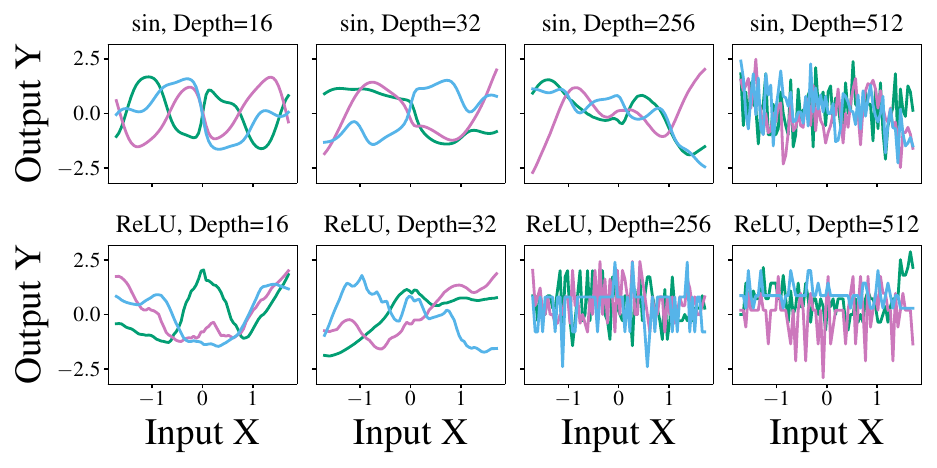}
  \end{subfigure}
  \hfill
  \begin{subfigure}[t]{0.48\textwidth}
  \centering
\includegraphics[width=0.85\linewidth]{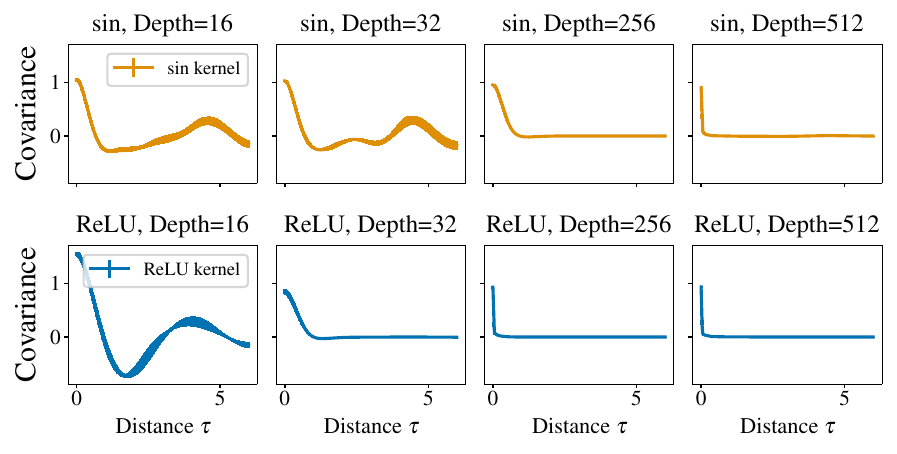}
  \end{subfigure}
  \caption{\textbf{Depth pathologies in randomly initialized neural networks.}
     Predictions of neural networks \textbf{(top)} with varying activations \textbf{(rows)} and depths \textbf{(columns)};
    mean and standard error of the covariance of learned kernels
    \textbf{(bottom)}.
    Greater depth results in kernels with shorter lengthscales, and this pathology emerges earlier in \glspl{relu-nn}.    
  }
  \label{fig:depthwidth}
\end{figure}

\subsection{Reproduction: Depth pathologies in randomly initialized NN}
\label{sec:random-init-prior}
Hyperparameter selection in \glspl{nn} is not always theoretically grounded.
Many recent studies thus characterize how hyperparameter choices affect the properties of \glspl{nn} at random initialization~\parencite{schoenholz2017deep,yang2019,Xiao2018DynamicalIA}.
Towards that end, recent work showed that increasing depth can actually induce pathologies in randomly initialized \glspl{nn} \parencite{Labatie2019CharacterizingWV, duvenaud2014pathologies}.
For example, \textcite{duvenaud2014pathologies} proved that increasing depth in a certain class of infinitely wide \glspl{nn} produces functions with ill-behaved derivatives.

We empirically study a similar pathology---quick variation in input space---that emerges in randomly initialized, finite-width, finite-depth \glspl{nn}.
To do this, we fit \gls{gp} surrogates to randomly initialized \gls{nn} ensembles with varying depths and activations.
If \glspl{nn} exhibit this pathology, the learned covariance will decay sharply with distance.

\paragraph{\Gls{nn} hyperparameters.}
We consider families of \glspl{nn} with \gls{sin} or \gls{relu} activations
and depths ranging from 16 to 512 layers.
From each family, we randomly initialize an ensemble of 50 \glspl{nn}, with 128 hidden units in each layer.
We randomly initialize \gls{nn} weights about zero with weight variance $\sigma_w^2=1.0$ and bias variance $\sigma_b^2=0.05$. 
\paragraph{\Gls{gp} surrogate.}
We learn an \gls{smk} kernel by optimizing \cref{eq:objective} separately for each ensemble,
running Adam for 750 iterations with $\eta = 0.1$.
We choose the kernel hyperparameters with the highest mean marginal likelihood among three random initializations.
To ensure our results are robust across an neural network ensemble, we consider the averaged learned kernel.
That is, if we have $n$ kernels, $k_1(\cdot), \ldots, k_n(\cdot)$, learned from $n$ different ensembles, we consider the average learned kernel $\bar{k}(\tau) = \frac{1}{n} \sum_{i=1}^{N} k_i(\tau)$.

\paragraph{Results.}
\cref{fig:depthwidth} plots the average learned kernels for \gls{nn} families with varying activation functions and depths, and the corresponding neural network predictions.
Across both activation functions, the learned kernels reveal a pathology:
For large depths, the covariance (\cref{fig:depthwidth}, bottom) decays sharply towards zero with distance.
The \gls{nn} predictions (\cref{fig:depthwidth}, top) explain this property:
At large depths, the \glspl{nn} vary quickly in the input domain, which leads the \gls{smk} to learn short lengthscales.
Interestingly, this pathology emerges at different depths for different activation functions.
We see \glspl{relu-nn} exhibit this pathology with 256 layers while \glspl{sin-nn} exhibit this pathology with 512 layers.

\subsection{Amortized influence estimation}
\begin{figure}[t]
  \centering
  {\includegraphics[scale=0.2]{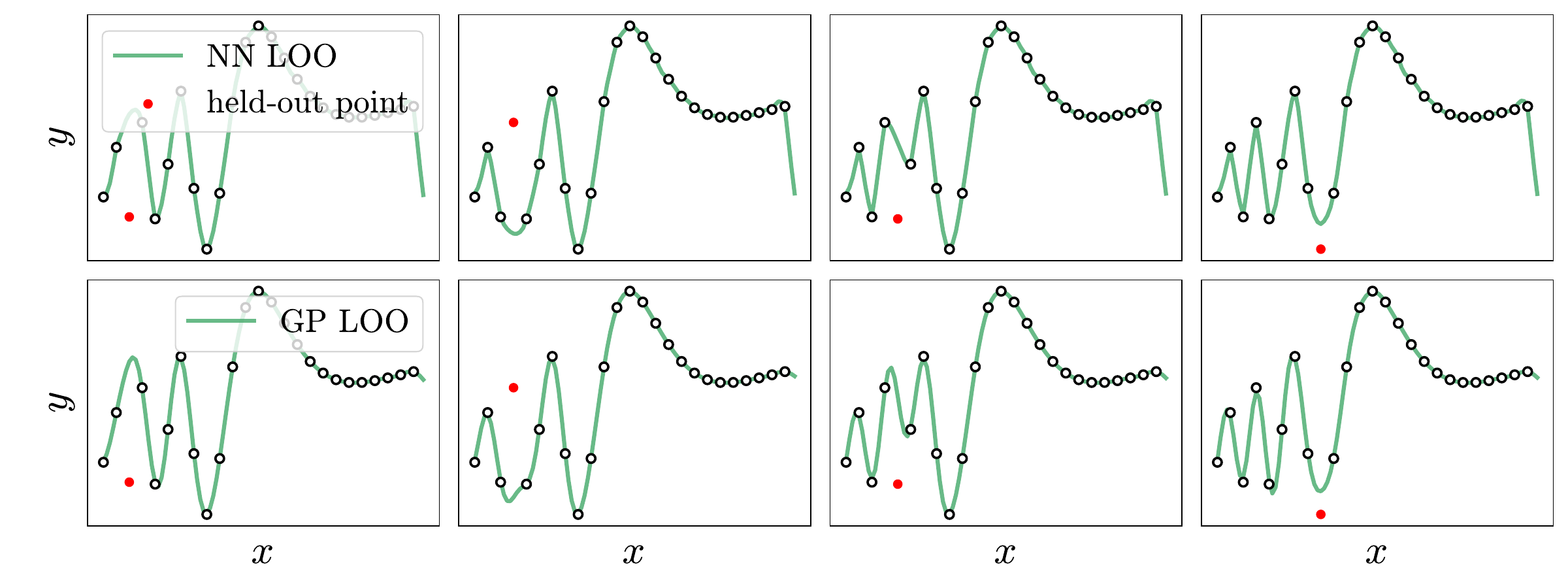}}
  \includegraphics[scale=0.23]{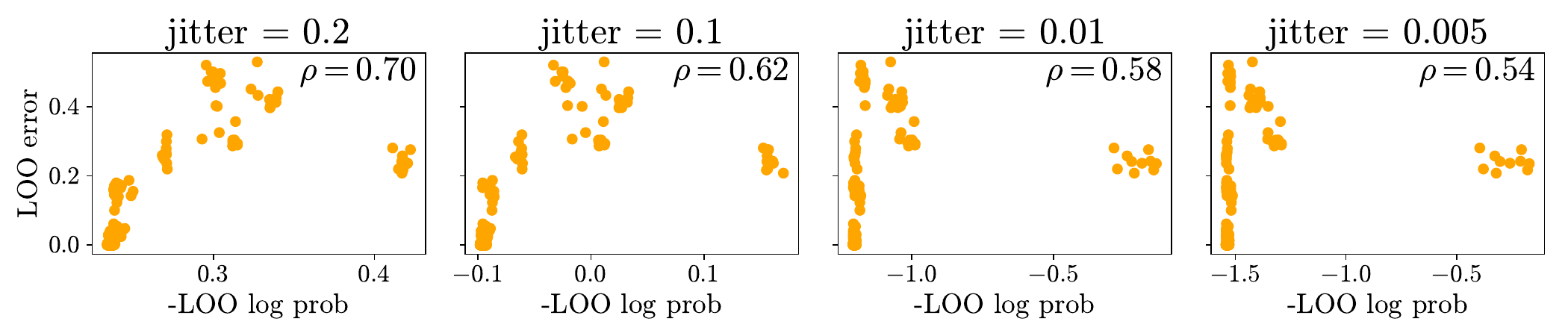}    
  \caption{
  \textbf{NN and GP surrogate leave-one-out predictions compared.}
  (\textbf{Top}) Neural network predictions obtained from re-training on target function with influential (red) training points removed. 
  (\textbf{Middle}) GP predictions on same datasets with kernel learned from trained NN predictions.
  Columns correspond to different held-out training points. 
  (\textbf{Bottom}) Correlation between LOO error and (negative) GP LOO predictive probability of held-out points; a lower negative LOO predictive probability means a less influential point.
  The GP and NN LOO predictions are consistent, and the LOO error positively correlates with the negative LOO predictive probability. 
 }
  \label{fig:loos_compared}
\end{figure}
\label{sec:influence-functions-label}
The primary computational bottleneck of fitting a GP is the $O(N^3)$ cost of Gram matrix inversion. 
However, once this matrix is inverted, certain computations that seem costly can actually be performed cheaply.
For example, the leave-one-out predictive (LOO) distributions can be computed analytically given the inverted Gram matrix and the training data~\citep{rasmussenbook}.
The LOO distributions characterize how GP predictions change after removing a training point.
In this section, we leverage this property of GPs to perform influence function analysis in the spirit of \textcite{koh2017understanding}. 
In particular, we use the LOO predictive distributions of a surrogate GP to detect influential points for \glspl{nn} with Fourier feature layers~\parencite{ffsletnns}, which enable \glspl{nn} to learn higher frequencies.
Similar architectures have achieved state-of-art performance in a number of important tasks in computer vision and computer graphics \citep{mildenhall2020nerf}.
We show that the surrogate LOO predictions are consistent with NN LOO predictions obtained from costly retraining.

\paragraph{\Gls{nn} hyperparameters.}
We train 10 NNs with 4 hidden layers of 1024 units, \gls{relu} activations \parencite{ffsletnns}.
The inputs to the NNs are passed through a single Fourier feature layer with parameter $p$ that controls the spectral bias of the NN. 
For details see \textcite{ffsletnns}.
We set $p=1$ and choose Fourier feature mappings with frequencies ranging from 1 to 4.
We train the network using full-batch gradient descent with \gls{sgd} and a learning rate of $\eta = 10^{-3}$ for 2000 iterations.

\paragraph{Target functions.}
We consider a target function inspired by one used in the leave-one-out GP literature \citep{fastcvgp2021}. 
The functional form is given by $f(x) = \sin(30(x-0.95)^4
)\cos(2(x-0.95)) + (x-0.95)/2
$.
This target function is non-stationary, with higher frequency behavior in the left half of its input domain and lower frequency behavior in the right.

\paragraph{\Gls{gp} surrogate.}
We fit a SMK with $Q = 10$ mixture components to trained neural network predictions at 300 input points evenly spaced between (0, 3). 
We initialize frequency hyperparameters by sampling uniformly from 0 to 20. 

We briefly overview leave-one-out cross-validation (LOO-CV) in the context of GPs.  
The GP LOO predictive probability when leaving out training example $i$ is
\begin{align}
    \log p(y_i | X_{-i}, \mathbf{y}_{-i}) = -\frac{1}{2} \log \sigma_i^2 - \frac{(y_i - \mu_i)^2}{2\sigma_i^2} - \frac{1}{2}\log2\pi   
  \label{eq:loo_equation}
\end{align}
The notation $\mathbf{y}_{-i}$ indicates all training targets \textit{besides} the $i$th target.
The LOO-CV predictive mean $\mu_i$ and variance $\sigma_i^2$ can be computed from the inverse of the Gram matrix: 
\begin{align}
 \mu_i = y_i - \frac{[K^{-1}y]_i}{[K^{-1}]_{ii}}, \quad \textrm{and} \quad \sigma_i^2 = \frac{1}{[K^{-1}]_{ii}}
\end{align}
Importantly, to compute these predictive means and variances, we can reuse the inverse of the Gram matrix computed to perform GP inference (\cref{eq:pp-mean,eq:marginal}).
In our analysis, we  will quantify the influence of a point $(x_i, y_i)$ using the predictive log probability $\log p(y_i | X_{-i}, \mathbf{y}_{-i})$.  
We can interpret points with lower log probability as more ``influential'' under the assumptions of a kernel. 

\paragraph{Results}
In the first two rows of Figure 4, we compare the surrogate GP LOO and NN LOO predictions for some of the most influential points. 
Indeed, the GP LOO predictions at the influential points are consistent with the NN LOO prediction. 
In the last row of Figure 4, we plot the (negative) LOO log probability at each input point against the actual LOO error at the training input. 
Each plot in the last row contains the data across all 10 \glspl{nn}.
The LOO error is computed as the difference between the NN LOO predictions and the predictions of the NN trained on the full dataset.
We validate this result across different levels of jitter added to the Gram matrix.
The Spearman's rank correlation coefficient $\rho$ shows a positive relationship between the negative LOO log probability at an input point and the LOO error at the training input.
Interestingly, the outliers on the right region of each subplot always correspond to the leftmost points in the training data.
This is consistent with previous work showing that the GP LOO can overestimate the influence of a boundary point  
\parencite{fastcvgp2021}.
\subsection{Predicting the generalization gap}
\label{sec:gen-gap}
\begin{figure}[t]
  \centering
  {\includegraphics[scale=0.75]{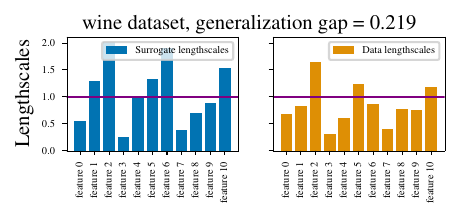}}
 
  {\includegraphics[scale=0.75]{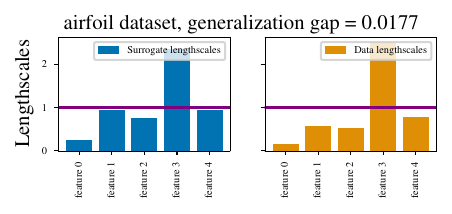}}
  \caption{
    \textbf{Qualitative connection between lengthscale profile discrepancy and generalization gap.}
    Each subfigure compares normalized lengthscales learned from neural network  predictions on validation set (\ie surrogate lengthscales) after training and normalized lengthscales learned from training data (\ie data lengthscales).
    A lengthscale greater than 1 indicates an ``unimportant'' feature.
    The title indicates the UCI dataset and generalization gap defined in \cref{fig:lengthscale_prof}.
    Data and surrogate lengthscales for some features are different (\eg features 1, 4, 6), reflected in a high generalization gap
    (\textbf{top}).
    Data and surrogate lengthscales for the same features are generally similar, reflected in a low generalization gap
    (\textbf{bottom}).
    This suggests a connection between the generalization gap and discrepancy between surrogate and data lengthscales.
  }
  \label{fig:lengthscale_qual}
\end{figure}
\begin{figure}[t]
  \centering
 {\includegraphics[trim=0 0 100 0,clip,,width=230pt]{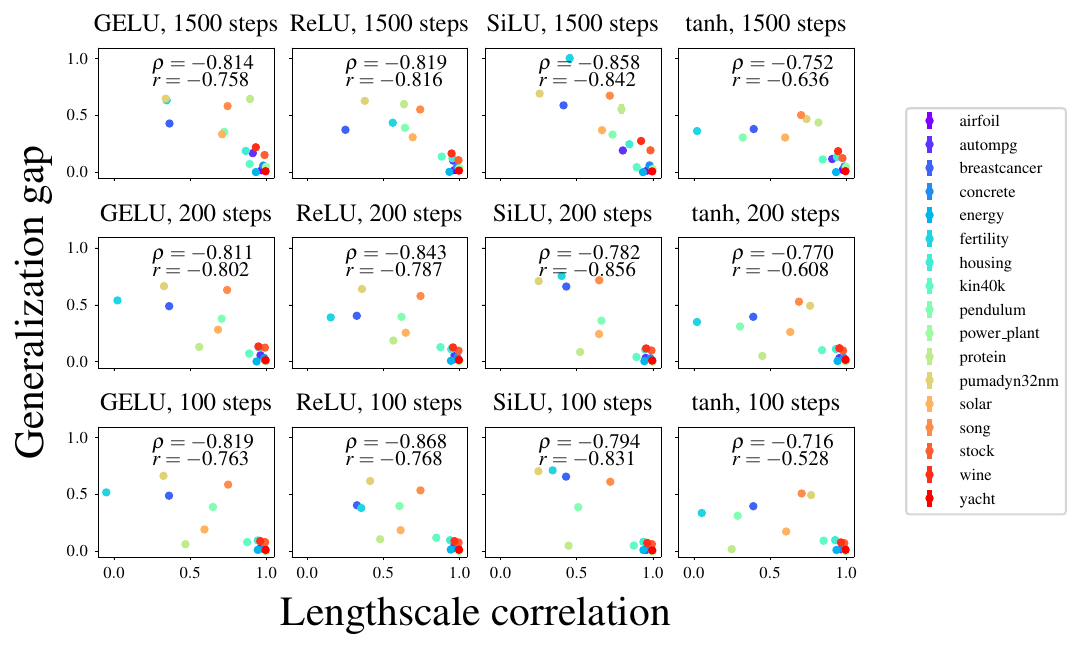}} 
  \caption{
    \textbf{Inverse relationship between generalization error and lengthscale correlation on UCI datasets.}
    Each point represents 
    the lengthscale correlation (between surrogate and data lengthscales) and 
    the average generalization gap for a neural network ensemble to which the surrogate model is fit,
    on a single UCI dataset; we also report the standard error of the generalization gap.
    Each panel corresponds to a particular neural family;
    see \cref{sec:gen-gap} for details about hyperparameters of these families, including architectures.
    Colors correspond to UCI datasets.
    Across datasets and architectures, a larger lengthscale correlation 
    (\ie higher similarity between the data and surrogate representations) 
    corresponds to a lower generalization gap 
    (\ie better extrapolation). 
  }
  \label{fig:lengthscale_prof}
\end{figure}

In these sections, we predict generalization using kernels learned from trained \glspl{nn}. 
To do this, we characterize trained \gls{nn} properties on the \textit{validation set} and compare these properties to the training data.
We focus on the validation set because it is more informative of extrapolation.
If the \gls{nn} extrapolates well, its predictions on the validation set should be ``similar'' in some sense to the dataset.
On the other hand, significant discrepancies could indicate poor extrapolation.
This intuition motivates us to compare a kernel from the training data with a surrogate kernel fit to \gls{nn} predictions on a validation set.
We find that lower similarity between these kernels correlates with a larger \textit{generalization gap} (\ie poorer extrapolation), defined as the difference between test error and training error \parencite[\eg][]{predictinggeneralization2020}.

\subsubsection{Initial demonstration}
\label{sec:gen-gap-small}

\paragraph{\Gls{nn} hyperparameters.}
We train ensembles of randomly initialized \glspl{nn} with \gls{silu}~\parencite{SILU}, \gls{gelu}~\parencite{Hendrycks2016GaussianEL}, \gls{relu}~\parencite{fukushima1975cognitron,nair2010rectified}, or \gls{tanh} activations, and two layers of 128 hidden units.
We use the LeCun normal initialization with a scale of 1.5 \parencite{LeCun2012EfficientB}.
We train 25 \glspl{nn} with full-batch gradient descent using \gls{adam} with a learning rate of $\eta = 0.003$.
We want to assess if our approach can distinguish between \glspl{nn} 
with similar training behavior but varying generalization performance,
so we train \glspl{nn} either for a maximum number of iterations, a hyperparameter,
or until training error reaches zero. 

\paragraph{Target functions.}
We consider a set of naturalistic regression tasks from the \gls{uci} dataset~\parencite{Dua:2019}, spanning a range of dataset sizes and input dimensions.
We split each of the datasets into a 72/8/20 train/validation/test split.
Both the data input and output are standardized by mean-centering and dividing by the standard deviation dimension-wise so that the target values and each dimension of the data input have near zero mean and unit variance.
We subsample 2,000 datapoints for datasets with more than 2,000 datapoints, as in \textcite{simpson2021kernel} and \textcite{liu2020ahgp}.

\paragraph{\Gls{gp} surrogate.}
We learn a \emph{data kernel} directly from the training dataset.
We also learn a \emph{surrogate kernel} from \gls{nn} predictions on the validation set.
In both cases, we use the \glsfirst{mk} since the \gls{smk} can struggle for higher-dimensional inputs.
We learn a separate lengthscale for each input dimension (\ie feature)  of the data.
We denote the lengthscales for a kernel as its \emph{lengthscale profile}.
We call the data kernel's lengthscales the \emph{data lengthscales} and the surrogate kernel's lengthscales the \textit{surrogate lengthscales}.
To quantify the mismatch between \gls{nn} validation predictions and the training data, we consider the \textit{correlation in lengthscale profiles across features}.
This is the correlation between the data and surrogate lengthscales.

\paragraph{Results.}
\cref{fig:lengthscale_qual} gives intuition for our more general result in \cref{fig:lengthscale_prof}.
For two \gls{uci} datasets, we compare the data lengthscales and the surrogate lengthscales for a two-layer \gls{gelu} \gls{nn}.
The vertical axis corresponds to (normalized) learned lengthscales for each input dimension.\footnote{
  For this visualization, we divide the learned lengthscale for each dimension by the difference between the maximum feature value and minimum feature value for each dimension.
  Therefore, a lengthscale that is much greater than 1 suggests that the \gls{nn} predictions do not vary much along that dimension.
}
When the generalization gap is small, the data kernel and surrogate kernel are similar; the same features have similar lengthscales (\cref{fig:lengthscale_qual}, bottom).
When the generalization gap is large, the data kernel and surrogate kernel have discrepancies.
For example, the surrogate lengthscales for features 1 and 6 are larger than 1, but the data lengthscales for feature 1 and 6 are smaller than 1 (\cref{fig:lengthscale_qual}, top).

In \cref{fig:lengthscale_prof}, we summarize our results across different architectures, datasets, and maximum training iterations.
We display the generalization gap against the correlation in lengthscale profiles across features.
We report the mean generalization gap along with its standard error.
The similarity in lengthscale profiles negatively correlates with generalization gap across a range of architectures and max iterations.
The Pearson correlation coefficients ($r$) range from \num[mode=text]{-0.856} to \num[mode=text]{-0.528}.
The Spearman correlation coefficients ($\rho$) range from \num[mode=text]{-0.868} to \num[mode=text]{-0.716}.
In Section A.2, we demonstrate these results are insensitive to outlier datasets.

\subsubsection{Larger-scale demonstration}
\label{sec:gen-gap-extension}
\begin{figure}[t]
  \centering
  {\includegraphics[width=240pt]{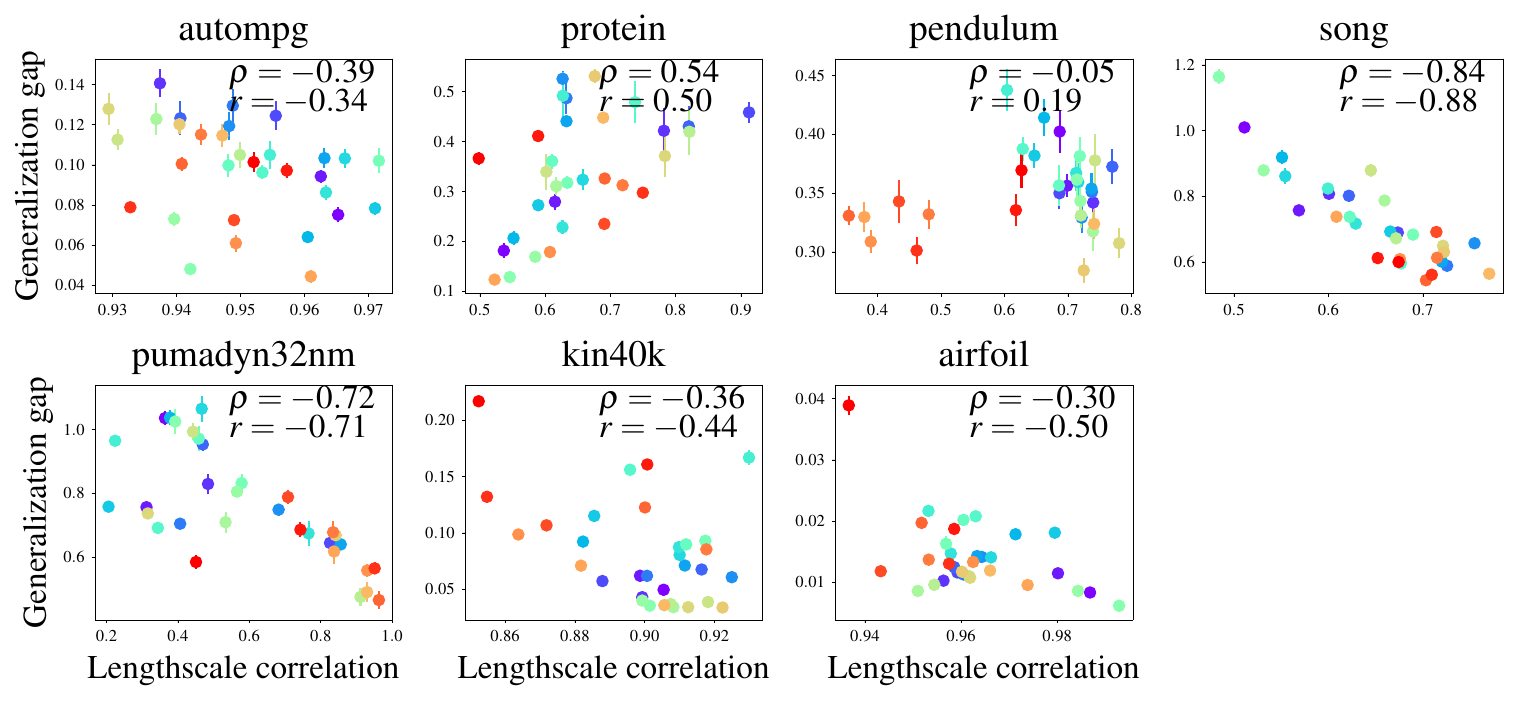}}
  \caption{
    \textbf{Inverse relationship between generalization error and lengthscale correlation on UCI datasets across hyperparameter sweep of neural networks.}
    Each point represents 
    the lengthscale correlation (between surrogate and data lengthscales) and 
    the generalization gap for a neural network ensemble to which the surrogate model is fit.
    Each panel corresponds to a particular UCI dataset;
    see \cref{sec:gen-gap-extension} for details about hyperparameters and architectures.
    In 5/7 datasets, a larger lengthscale correlation 
    (\ie higher similarity between the data and surrogate representations) 
    corresponds to a lower generalization gap 
    (\ie better extrapolation). 
  }
  \label{fig:lengthscale_prof_hparam_sweep}
\end{figure}
We now extend the analysis in the previous section to a larger hyperparameter sweep, more closely emulating the analysis a practitioner would use in selecting a model for a dataset.  

\paragraph{\Gls{nn} hyperparameters.}
We train ensembles of randomly initialized \glspl{nn} with four activations 
(\gls{silu}, \gls{gelu}, \gls{relu}, or \gls{tanh}), 
two depths (2, 4), two widths (32, 64), and two learning rates with Adam (0.03, 0.003),
generating 32 hyperparameter combinations.
We use the LeCun normal initialization with a scale of 1.5.
We train 50 \glspl{nn} for 500 iterations with full-batch gradient descent using \gls{adam}.

\begin{figure}[t]
  \centering
{\includegraphics[width=240pt]{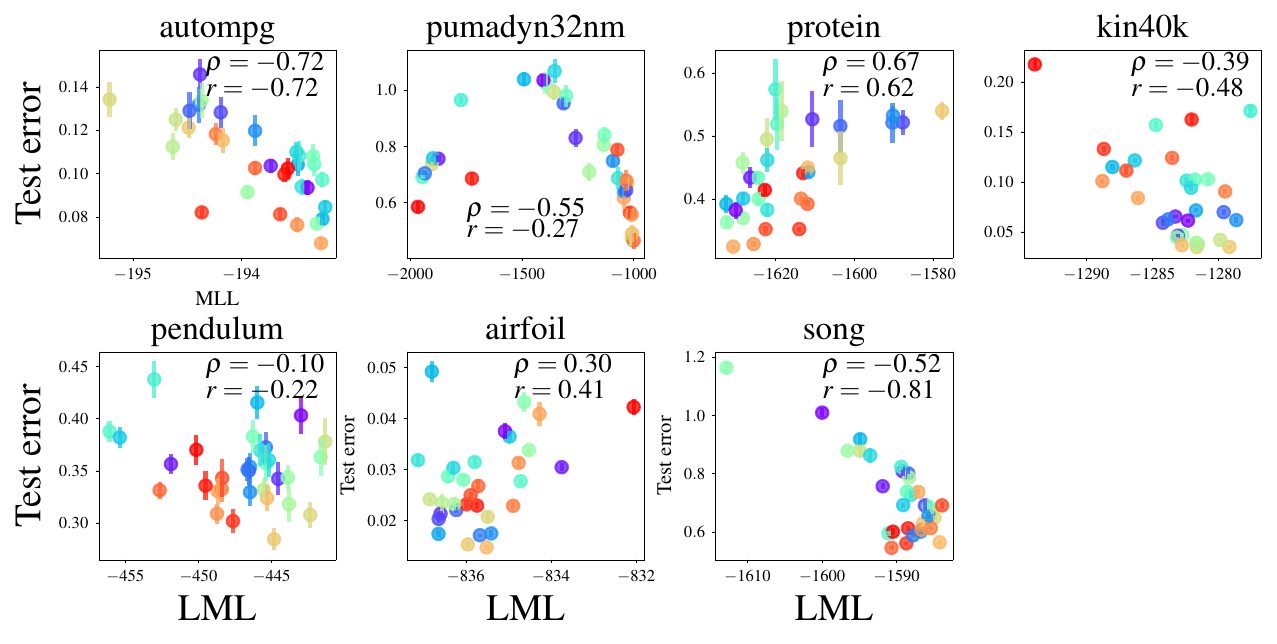}}   
  \caption{
    \textbf{Inverse relationship between test error and marginal likelihood on UCI datasets}
    Each point represents the marginal likelihood of the training data (using the surrogate kernel for a neural network family) and the test error for a neural network ensemble to which the surrogate model is fit.
    Each panel corresponds to a particular UCI dataset;
    see \cref{sec:gen-gap-extension} for details about hyperparameters of these families, including architectures.
    Each color corresponds to a particular neural network ensemble.
    In 5/7 datasets, a larger marginal likelihood correlates with lower test error. 
  }
  \label{fig:lml_hparam_sweep}
\end{figure}
\paragraph{Target functions.}
We consider a subset of UCI datasets that satisfy two criteria: more than 2 unique feature values for each dimension, and a range in lengthscale correlations between neural network and datasets greater than 0.025;
the first condition ensures that we can learn sensible lengthscales for the dataset, while
the second ensures that there is meaningful variation in \gls{nn} behavior within a dataset.
As in \cref{sec:gen-gap-small}, 
we split each of the datasets into a 72/8/20 train/validation/test split, subsample 2,000 datapoints if necessary, and standardize data inputs and outputs.

\paragraph{\Gls{gp} surrogate.}
In addition to the correlation in lengthscale profiles across features from \cref{sec:gen-gap-small}, 
we consider the marginal likelihood of the training dataset under the kernel learned for each neural network family.

\paragraph{Results.}
\Cref{fig:lengthscale_prof_hparam_sweep} plots the average generalization gap and  standard error against the correlation in lengthscale profiles across features.
In contrast to \cref{fig:lengthscale_prof}, each panel corresponds to a particular dataset and each data point corresponds to a particular neural network family with a set of hyperparameters described above.  The majority of datasets exhibit a negative relationship between correlation  in lengthscale profiles and generalization gap. There are two exceptions: pendulum and protein.
For the pendulum dataset, the positive relationship is driven by neural networks with tanh activations (the dots in different shades of red). The Matern kernel may struggle to model the tanh networks, consistent with \cref{fig:lengthscale_prof} where the correlations were consistently lower for the tanh networks.
For the protein dataset, the positive relationship may be due to challenges in fitting GPs to this dataset; recent work showed that exact GP regression on a training dataset (without any subsampling) from the protein dataset attains high test RMSE \parencite{Wang2019ExactGP}.

In \cref{fig:lml_hparam_sweep},
we plot the test error against the marginal likelihood of the training data under the various surrogate kernels for each neural network ensemble.
That is, each point corresponds to the marginal likelihood of the training data using the surrogate kernel learned from each neural network family.
We find the marginal likelihood correlates with test error on several of the datasets (each dataset is indicated by the subplot title).
Consistent with previous work \parencite{lotfi2022bayesian}, we find that this result is sensitive to the jitter (which we set to 0.5). 
In Section A.1 of the supplement, we sanity check this analysis on synthetic datasets. 

\section{Discussion}
In this work, we illustrated the potential of empirically characterizing neural networks with Gaussian process surrogates.
We captured the spectral bias of deep rectifier networks by examining how the surrogate kernel evolves during training (\cref{sec:spectral-bias}).
We captured pathologies that emerge in randomly initialized neural networks by examining the learned kernels of neural networks with varying depths (\cref{sec:random-init-prior}). 
We further demonstrated that Gaussian process surrogates can be used to identify influential datapoints (\cref{sec:influence-functions-label})
and predict neural network generalization (\cref{sec:gen-gap}).
Taken together, these results suggest that Gaussian process surrogates may be a valuable empirical tool for investigating deep learning,
and future work could aim to use this framework to complement existing approaches to interpretability~\parencite[\eg][]{Ribeiro2016WhySI}
and examining extrapolation behavior~\parencite[\eg][]{xu2021how}.

We note some limitations.
First, though the framework is, in principle, applicable to broader settings, we restricted this first exploration to regression tasks and feed-forward neural network architectures.
A broader study of more architectures and tasks would be challenging due to the need to scale Gaussian processes but potentially rewarding, as characterizing properties of neural networks is a open problem with far-reaching implications~\parencite{Sejnowski2020TheUE}.
To tackle this, we could leverage techniques such as inducing points ~\parencite[\eg][]{pmlr-v5-titsias09a}.
Another limitation is that the data generation process involves training many networks.
We could leverage recent work on fast neural network training \parencite{leclerc2022ffcv}.

Second, we learn point estimates of kernel hyperparameters~\parencite[type II maximum likelihood;][]{Gelman2013BayesianDA}.
Although this is standard, we could infer the posterior over hyperparameters using \gls{mcmc} or variational inference \parencite{pmlr-v118-lalchand20a, slicesamplegp} to perform a fully Bayesian analysis.
We could also explore a richer set of kernels \parencite{pmlr-v28-duvenaud13}.
These are exciting avenues for future work.



\begin{acknowledgements} 
We thank Jake Snell for helpful discussions. This work was supported by ONR grant number N00014-18-1-2873.

\end{acknowledgements}
\bibliography{uai2023-template}

\title{Gaussian Process Surrogate Models for Neural Networks \\(Supplementary Material)}

%
%

  
\onecolumn 
\maketitle



\appendix
\section{Additional experimental results}
\subsection{Ranking NN generalization with the GP marginal likelihood}
In previous sections, we demonstrated that \gls{gp} surrogate models could yield insight into \gls{nn} behavior.
The benefits of \glspl{gp} extend beyond this.
Since the \gls{gp} marginal likelihood has a closed form expression, many have advocated for using the marginal likelihood in model selection and as an indicator of expected generalization performance \parencite{Mackay1992APB}.
In this section, we leverage the learned \gls{gp} surrogate to \textit{rank NNs by their generalization error with the GP marginal likelihood}.
In particular, we learn \gls{gp} surrogates from different \glspl{nn} at random initialization, and
we then study if the marginal likelihood of the surrogates can rank the \glspl{nn} by test error after training.
In the following experiments with varying classes of \gls{nn} families, we find that we can indeed predict test error using the marginal likelihood  of the training set under the learned surrogate \gls{gp}.

\subsubsection{The idealized case: Large-width \glspl{nn}}
\label{sec:gen-rank-large-width}

Before we consider arbitrary \gls{nn} families, we check that the marginal likelihood is predictive in an idealized setting.
In particular, we consider large-width \glspl{nn} whose infinite-width analogs are equivalent to \gls{gp}s \parencite{leedeep}.  %
If the marginal likelihood is not predictive in this case in which the kernel function can be analytically determined, it is unlikely to be useful in a general setting where the kernel is learned and \glspl{gp} approximate \glspl{nn} priors but are not equivalent.

\paragraph{\Gls{nn} hyperparameters.}
We consider \glspl{nn} with \gls{sin} or \gls{erf}\footnote{
Here, \gls{erf} is defined as $a \,\erf(bx)+c$, where $\erf(x) = \frac {2}{\sqrt {\pi }} \int _{0}^{x}e^{-t^{2}}\,\mathrm{d}t$.
}
activations and 2 hidden layers of 1024 units each.
We randomly initialize the weights about zero with weight variance $\sigma_w^2=1.5$ and bias variance $\sigma_b^2=0.05$. 
We train an ensemble of 50 randomly initialized \glspl{nn} from each family using full-batch (vanilla) gradient descent with learning rates of $\eta \in \{0.01, 0.1\}$.

\begin{figure}
  \centering
  \adjincludegraphics
  [Clip={.0\width} {.0\height} {0.48\width} {.0\height},
    height=110pt]
  {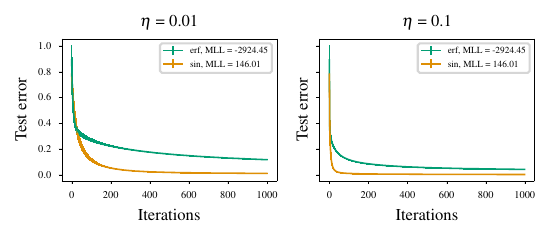}
  \adjincludegraphics
  [Clip={.58\width} {.0\height} {0.0\width} {.0\height},
    height=110pt]
  {figures/plots/predictability_sanity.pdf}
  \caption{
    \textbf{Ranking generalization from MLL in large-width NNs.}
    Mean and standard error of the test MSE
    of large-width
    sinusoidal and \acrshort{erf} \glspl{nn}
    trained with 
    learning rates
    $\eta = 0.01$ \textbf{(left)} and
     $\eta = 0.1$ \textbf{(right)}
    on the target function of \cref{sec:gen-rank-large-width}.
    The \acrshort{mll} of the target function under the surrogate model corresponding to the limiting kernel for each model family is shown in the legend. 
    Consistent with expectations, the model family whose surrogate assigns higher MLL to the target function achieves lower test error for both values of $\eta$.
  }
  \label{fig:predictabilitysanity}
\end{figure}
\paragraph{Target function.}
The target function is $ \sin(0.5x)$.

\paragraph{\Gls{gp} surrogate.}
We do not learn a kernel from \gls{nn} predictions as in previous sections.
Instead, we use the kernels corresponding to the infinite width analogs of the \glspl{nn} using the neural-tangents package
\parencite{neuraltangents2020}.

\paragraph{Results.}
\cref{fig:predictabilitysanity} compares the performance of these \gls{nn} families along with the marginal likelihood of the target function under the surrogate model.
The performance (\gls{mse} on the test set) is averaged across each ensemble of \glspl{nn}.
The \gls{mll} of the target function is higher for the better-performing \gls{nn} family.

\subsubsection{Small width neural networks and learning the kernel}
\label{sec:gen-rank-small-width}

In the previous experiment, we showed that the marginal likelihood could be predictive when we consider large-width \glspl{nn} and when we use a corresponding, analytically derived kernel.
Is the marginal likelihood predictive when we consider
smaller-width \glspl{nn} and when we learn the kernel empirically?

\paragraph{\Gls{nn} hyperparameters.}
We consider ensembles of width 16, depth 4 \glspl{nn} from two families: \glspl{nn} with \gls{sin} activations and \glspl{nn} with \gls{relu} activations.
We randomly initialize weights about zero with weight variance $\sigma_w^2=1.5$ and bias variance $\sigma_b^2=0.05$. 
We train an ensemble of 50 randomly initialized \glspl{nn} from each family on the target functions using full-batch gradient descent with a learning rate of $\eta = 0.1$.

\paragraph{Target function.}
The target function families mirror the \gls{nn} model families: We collect predictions from randomly initialized, width 16, depth 4 \glspl{nn} with \gls{sin} or \gls{relu} activations.
These target functions are a useful sanity check, as the inductive biases of the model families are perfectly suited for a target function family.

\begin{figure}
  \centering
  \begin{subfigure}{0.5\textwidth}
    \centering
    {
      \includegraphics[scale=0.9]{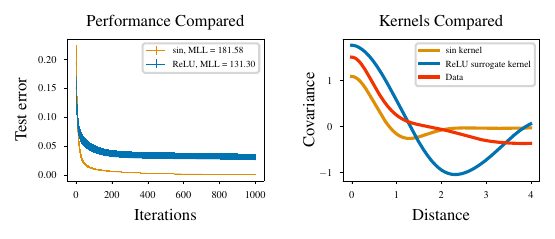}
    }
    \label{fig:predictabilityexp1}
  \end{subfigure}
  \begin{subfigure}{0.5\textwidth}
    \centering
    {
      \includegraphics[scale=0.9]{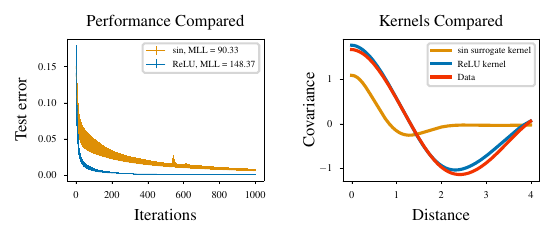}
    }
    \label{fig:predictabilityexp2}
  \end{subfigure}

  \caption{
    \textbf{Ranking generalization from MLL in small-width NNs.}
    Mean and standard error of test MSE 
    \textbf{(left)}
    of small-width sinusoidal and rectifier \gls{nn} ensembles on \gls{sin} \textbf{(top)} and \gls{relu} \textbf{(bottom)} target function families,
    with the target function MLL under the surrogate learned from each model family in the legend.
    Covariance \textbf{(right)}
    of surrogate kernels alongside
    data kernels learned from the \gls{sin} \textbf{(top)} and \gls{relu} \textbf{(bottom)} target function families.
    Even in the small-width regime and when the kernel is learned, the model family whose surrogate assigns a higher MLL to the target function attains lower error
    \textbf{(left)};
    the surrogate kernel learned from the better-performing model family better matches the data kernel
    \textbf{(right)}.
  }
  \label{fig:predictabilityexp12}
\end{figure}

\paragraph{\Gls{gp} surrogate.}
For each ensemble, we learn the hyperparameters of an \gls{smk} with $Q=5$ mixture components by optimizing the marginal likelihood across the ensemble.
To optimize, we randomly initialize the kernel hyperparameters and run Adam for 250 iterations with a learning rate of $\eta = 0.1$.
We initialize the frequency parameters by sampling from a uniform distribution whose upper limit is the Nyquist frequency.
We choose the kernel hyperparameters with the highest objective value across three random initializations.

\paragraph{Results.}
In \cref{fig:predictabilityexp12}, we compare the performances of the two \gls{nn} families on the two target function families.
We also display the kernels learned
from \gls{nn} behavior (\emph{sin surrogate kernel} or \emph{\gls{relu} surrogate kernel})
and
learned from the target function family (\emph{data kernel}) directly.
Across both experiments, the \gls{mll} averaged across the target function family of the better-performing \gls{nn} family is higher.
In general, the structure of a learned kernel reflects the properties of the learned \gls{gp} prior,
and so we can compare kernels to assess similarity between target function and \gls{nn} families.
We see that the data kernel provides a better qualitative match to the kernel of the better-performing model family.

\subsubsection{Systematic study of various learning rates and architectures}
\label{sec:gen-rank-systematic}
\begin{figure}[t]
  \centering
  {
    \includegraphics[scale=1.0]{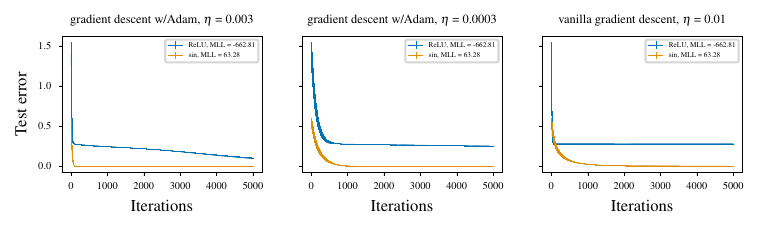}

    \includegraphics[scale=1.0]{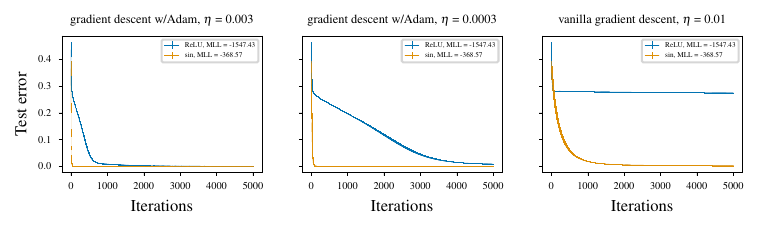}
  }
\caption{\textbf{Ranking generalization performance from MLL across different learning algorithms and architectures.}
Each panel displays mean and standard error of test MSE of an NN family trained on the target function $\sin(0.5x)$ with noise; legend displays \acrshort{mll} of the training data under the surrogate for one of two NN families:
1-layer (256 hidden units) sinusoidal or rectifier \glspl{nn} \textbf{(top)});
3-layer (256 hidden units) sinusoidal or rectifier \glspl{nn} \textbf{(bottom)}.
\Glspl{nn} are trained with batch gradient descent with Adam (learning rates $\eta= 0.003$, $\eta = 0.0003$) or vanilla batch gradient descent ($\eta = 0.01)$.
Across architectures and learning algorithms, the \gls{nn} family whose surrogate assigns higher MLL to the target function achieves lower test error.
}
\label{fig:predictability_panel}
\end{figure}

In this last experiment on ranking generalization performance, we establish that Gaussian process surrogates reliably rank performance across a range of learning rates and gradient descent algorithms.

\paragraph{\Gls{nn} hyperparameters.}
We consider ensembles of randomly initialized \glspl{nn} with \gls{sin} or \gls{relu} activations and 1 or 3 hidden layers with 256 hidden units in each layer.
We randomly initialize the weights about zero with weight variance $\sigma_w^2=1.5$ and bias variance $\sigma_b^2=0.05$. 
We train 50 randomly initialized \glspl{nn} from each family using either vanilla full-batch gradient descent with a constant learning rate of $\eta = 0.01$, or Adam~\parencite{kingma:adam} using learning rates of $\eta \in \{0.0003, 0.003\}$.

\paragraph{Target function.}
We consider a target function of $\sin(0.5x)$.
\paragraph{\Gls{gp} surrogate.}
For each ensemble, we learn the hyperparameters of an \gls{smk} with $Q=5$ mixture components by optimizing  the marginal likelihood across the ensemble.
To optimize, we randomly initialize the kernel hyperparameters and run \gls{adam} for 250 iterations with a learning rate of $\eta = 0.1$.
We choose the kernel hyperparameters with the highest objective value across three random initializations.
To randomly initialize the frequency parameters, we uniformly sample from the real-valued interval $(0, 25]$.

\paragraph{Results.}
In \cref{fig:predictability_panel},  we find that the marginal likelihood of the better-performing \gls{nn} family is higher.
The marginal likelihood depends on the diagonal noise $\sigma_n^2$ added to the Gram matrix.
We find that our result are robust across three levels of this diagonal noise ($10^{-3}, 10^{-4}, 10^{-5}$).
These results suggest we can rank these \gls{nn} families when they are not in the asymptotic regime and when we learn the kernel, in contrast to \cref{sec:gen-rank-large-width}, as well as when \emph{a priori} no model family should perform better, unlike \cref{sec:gen-rank-small-width}.

\begin{figure}[ht!]
  \centering
  {\includegraphics{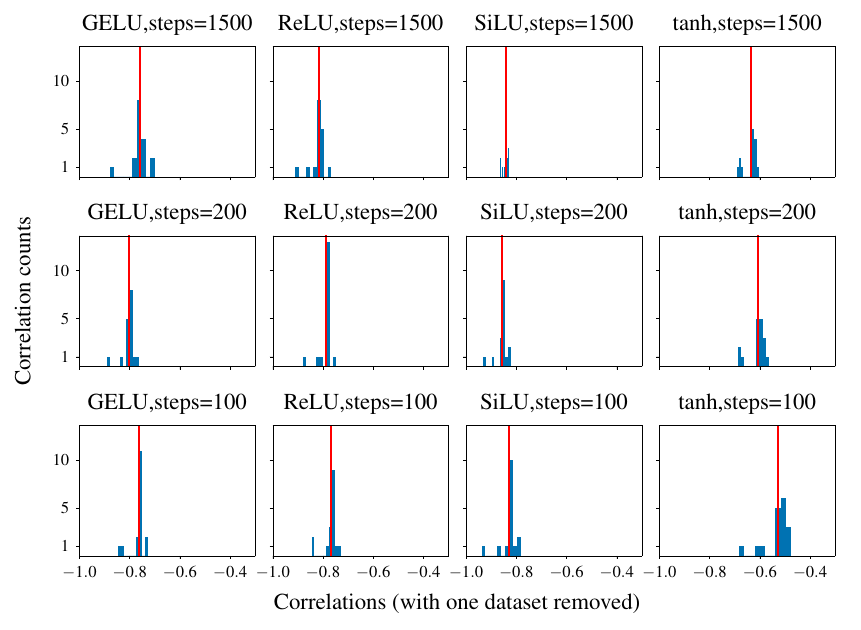}}
  \caption{
  \textbf{Sensitivity analysis of generalization gap and lengthscale profile relationship.}
  Each panel a histogram and mean (red line) of correlations obtained by recomputing the correlation between lengthscale profile correlation and generalization gap after removing each UCI dataset.
  Across datasets and architectures, even when a single dataset is removed, there remains an negative correlation between generalization gap and lenthscale profile correlation.
  Therefore, the inverse relationship between generalization gap and lengthscale profile correlation demonstrated in 
  Section 4.4.1 is robust to outlier datasets.  
 }
  \label{fig:corr_sensitivity}
\end{figure}

\subsection{Correlation sensitivity}
\label{sec:corr-sensitivity}
We present some additional results to supplement our analysis from Section 4.4.1 where we demonstrated that discrepancy in lengthscale profiles between data and neural network predicts the generalization gap.
Correlation can be sensitive to outliers.
Does any single dataset account for the negative correlations?
To answer this, we characterize how the correlation changes as a result of dropping each dataset.
Specifically, for each UCI dataset, we remove that dataset and then compute the correlation between lengthscale profile correlation and generalization gap for the remaining datasets.
We plot the resulting distribution of correlations in \cref{fig:corr_sensitivity}.
We find there is a tight spread around the correlation computed from all the UCI datasets.
Importantly, when we remove any UCI dataset, we still see moderate to high negative correlations between lengthscale profile correlation and generalization gap.

\begin{figure}[ht!]
  \centering
  {\includegraphics[scale=0.5]{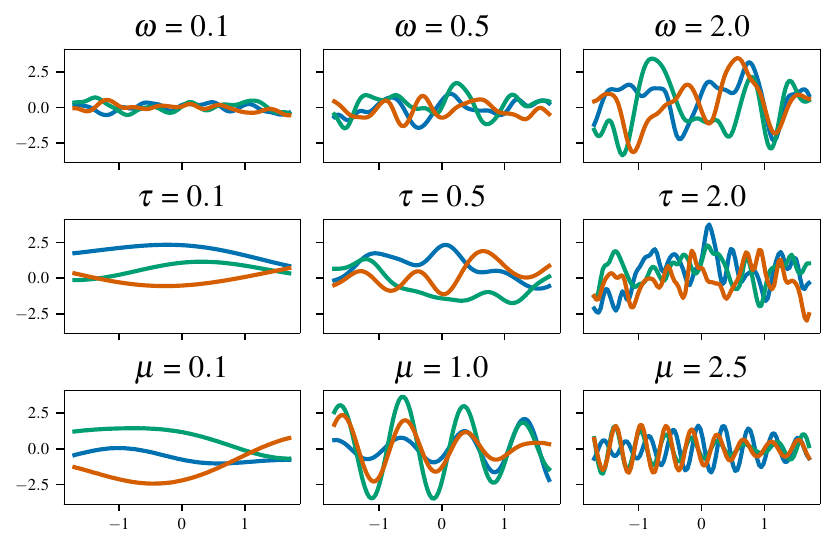}}
  {\includegraphics[scale = 0.5]{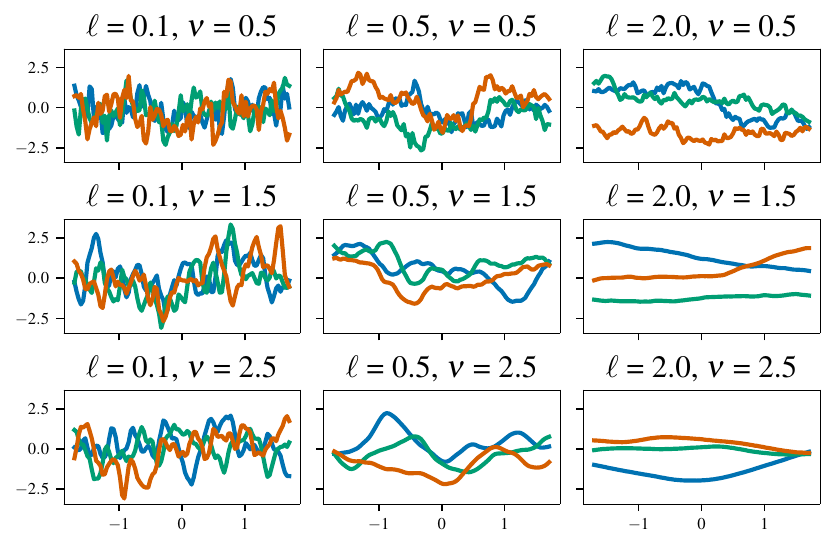}}
  \caption{
   \textbf{Illustrating the effect of \gls{gp} kernel hyperparameters on the \gls{gp} prior.}
(\textbf{Left}) Samples from a GP prior with \gls{smk} with varying mixture weights $\omega$, mixture scale $\tau$, and mixture means $\mu$.
(\textbf{Right}) Samples from a GP prior with Matern kernel with varying $\nu$ and $\ell$ (lengthscale).
\gls{gp}s are flexible models whose properties can be controlled through hyperparameters.
}
  \label{fig:gp_examples}
\end{figure}

\subsection{Properties of the Spectral Mixture Kernel and the Matern Kernel}
\label{sec:app-kernels}

We describe how the various hyperparameters of the \acrshort{smk} and \acrshort{mk} kernel affect the \gls{gp} prior.
We begin with the spectral mixture kernel.
The mixture weights $w$ are signal variances and control the scale of the function values.
The mixture means ($\mu$) encode periodic behavior.
The variances ($\tau$) are (inverse) lengthscales, which control the smoothness.
The (ARD) \gls{mk} kernel has lengthscales $\theta$, which controls the smoothness of the function with respect to each dimension.
$\nu$ is another hyperparameter that also modulates smoothness, and the Matern covariance function admits a simple expression when $\nu$ is a half-integer.
$\nu = 2.5$ corresponds to twice differentiable functions and $\nu = 1.5$ corresponds to once differentiable functions.

In \cref{fig:gp_examples}, we vary the hyperparameters of the \gls{smk} ($w, \mu, \tau$)and Matern kernels ($\nu, \theta$) and illustrate how they impact the prior over functions.

\clearpage


\end{document}